# Boosting $k$-NN for categorization of natural scenes

Paolo Piro · Richard Nock · Frank Nielsen · Michel Barlaud

**Abstract** The $k$-nearest neighbors ($k$-NN) classification rule has proven extremely successful in countless many computer vision applications. For example, image categorization often relies on uniform voting among the nearest prototypes in the space of descriptors. In spite of its good generalization properties and its natural extension to multi-class problems, the classic $k$-NN rule suffers from high variance when dealing with sparse prototype datasets in high dimensions. A few techniques have been proposed in order to improve $k$-NN classification, which rely on either deforming the nearest neighborhood relationship by learning a distance function or modifying the input space by means of subspace selection.

In this paper, we propose a novel boosting algorithm, called UNN (Universal Nearest Neighbors), which induces *leveraged* $k$-NN, thus generalizing the classic $k$-NN rule. Our approach consists in redefining the voting rule as a strong classifier that linearly combines predictions from the $k$ closest prototypes. Therefore, the $k$ nearest neighbors examples act as weak classifiers and their weights, called *leveraging coefficients*, are learned by UNN so as to minimize a *surrogate risk*, which upper bounds the empirical misclassification rate over training data. A major feature of UNN is the ability to learn which prototypes are the most relevant for a given class, thus allowing one for effective data reduction by filtering the training data.

Experimental results on the synthetic two-class dataset of Ripley show that such a filtering strategy is able to reject "noisy" prototypes, and yields a classification error close to the optimal Bayes error. We carried out image categorization experiments on a database containing eight classes of natural scenes. We show that our method outperforms significantly the classic $k$-NN classification, while enabling significant reduction of the computational cost by means of data filtering.

# 1 Introduction

## 1.1 Generic visual categorization

In this paper, we address the problem of generic visual categorization. This is a relevant task in computer vision, which aims at automatically classifying images into a discrete set of categories, such as *indoor* vs *outdoor*, *beaches* vs *mountains*, *churches* vs *towers*. Generic categorization is distinct from object and scene recognition, which are classification tasks concerning particular instances of objects or scenes (e.g. *Notre Dame Cathedral* vs *St. Peter's Basilic*). It is also distinct from other related computer vision tasks, such as content-based image retrieval (that aims at finding images from a database, which are semantically related or visually similar to a given query image) and object detection (which requires to find both the presence and the position of a target object in an image, e.g. person detection).

P. Piro · M. Barlaud
University of Nice-Sophia Antipolis / CNRS, 2000 route des Lucioles - 06903 Sophia Antipolis Cedex, France

P. Piro E-mail: piro@i3s.unice.fr · R. Nock
CEREGMIA Department, University of Antilles-Guyane, Martinique, France
E-mail: rnock@martinique.univ-ag.fr

F. Nielsen
Department of Fundamental Research, Sony Computer Science Laboratories, Inc., Tokyo, Japan
E-mail: nielsen@lix.polytechnique.fr

F. Nielsen
LIX Department, Ecole Polytechnique, Palaiseau, France
E-mail: nielsen@lix.polytechnique.fr

M. Barlaud
E-mail: barlaud@i3s.unice.fr

Automatic categorization of generic scenes is still a challenging task, due to the huge number of natural categories that should be considered in general. In addition, natural image categories may exhibit high inter-class variability (i.e., visually different images may belong to the same category) and low inter-class variability (i.e., distinct categories may contain visually similar images).

Classifying images requires a reliable description of the content relevant for an application (e.g., location and shape of specific objects or overall scene appearance). Examples of suitable image descriptors for categorization purposes are Gist, i.e. global image features representing the overall scene (Oliva and Torralba, 2001), and SIFT descriptors, i.e. descriptors of local features extracted at salient patches (Lowe, 2004).

Gist descriptor is based on the so-called "spatial envelope" (Oliva and Torralba, 2001), which is a very effective low dimensional representation of the overall scene based on spectral information. Such a representation bypasses segmentation, extraction of keypoints and processing of individual objects and regions, thus enabling a compact global description of images. Gist descriptors have been successfully used for categorizing locations and environments, showing their ability to provide relevant priors for more specific tasks, like object recognition and detection (Rubin et al, 2003).

## 1.2 $k$-NN classification

Apart from the descriptors used to compactly represent images, most image categorization methods rely on supervised learning techniques for exploiting information about known samples when classifying an unlabeled sample. Among these techniques, $k$-NN classification has proven successful, thanks to its easy implementation and its good generalization properties (Shakhnarovich et al, 2006). Indeed, the $k$-NN rule does not require explicit construction of the feature space and is naturally adapted to multi-class problems. Moreover, from the theoretical point of view, $k$-NN classification provably tends to the Bayes optimal when increasing the sample size. Although such advantages make $k$-NN classification very attractive to practitioners, it is an algorithmic challenge to speed-up $k$-NN queries and design schemes that scale-up well with large dimensional datasets (Shakhnarovich et al, 2006). Moreover, it is yet another challenge to reduce the misclassification rate of the $k$-NN rule, usually tackled by data reduction techniques (Hart, 1968).

In a number of works, the classification problem has been reduced to tracking ill-defined categories of neighbors, interpreted as "noisy" (Brighton and Mellish, 2002). Most of these recent techniques are in fact partial solutions to a larger problem related to nearest neighbors' error, which does not have to be the discrete prediction of labels, but rather a continuous estimation of class membership probabilities (Holmes and Adams, 2003). This problem has been reformulated by Marin et al (2009) as a strong advocacy for the formal transposition of *boosting* to nearest neighbors classification. Such a formalization is challenging as nearest neighbors rules are indeed not *induced*, whereas all formal boosting algorithms induce so-called *strong* classifiers by combining *weak* classifiers (also induced, say by decision stumps).

A survey of the literature shows that at least four different categories of approaches have been proposed in order to improve $k$-NN classification:

- learning local or global adaptive distance metric;
- embedding data in the feature space (kernel nearest neighbors);
- distance-weighted and difference-weighted nearest neighbors;
- boosting nearest neighbors.

The earliest approaches to generalizing the $k$-NN classification rule relied on learning an adaptive distance metric from training data. Refer to the seminal work of Fukunaga and Flick (1984) who presented an optimal global metric for $k$-NN. An analogous approach was later adopted by Hastie and Tibshirani (1996), who carried out linear discriminant analysis to adaptively deform the distance metric. Recently, Paredes (2006) has proposed a method for learning a weighted distance, where weights can be either global (i.e., only depending on classes and features) or local (i.e., depending on each individual prototype as well).

Other more recent techniques apply the nearest neighbors rule to data embedded in a high-dimensional feature space, following the kernel trick approach of support vector machines. For example, Yu et al (2002) have proposed a straightforward adaptation of the kernel mapping to the nearest neighbors rule, which yields significant improvement in terms of classification accuracy. In the context of vision, a successful technique has been proposed by Zhang et al (2006), which involves a "refinement" step at classification time, without relying on explicitely learning the distance metric. This method trains a local support vector machine on nearest neighbors of a given query, thus limiting the most expensive computations to a reduced subset of prototypes.

Another class of $k$-NN methods rely on weighting nearest neighbors votes based on their distances to the query sample (Dudani, 1976). Recently, Zuo et al (2008) have proposed a similar weighting approach, where the nearest neighbors are weighted based on their vector difference to the query. Such a difference-weight assignment is defined as a constrained optimization problem of sample reconstruction from its neighborhood. The same authors have proposed a kernel-based non-linear version of this algorithm as well.

Finally, only very few work have proposed the use of boosting techniques for $k$-NN classification. For instance, Amores et al (2006) use AdaBoost for learning a distance function to be used for $k$-NN search. On the other hand, García-Pedrajas and Ortiz-Boyer (2009) adopt the boosting approach in a non-conventional way. At each iteration a different $k$-NN classifier is trained over a modified input space. Namely, the authors propose two variants of the method, depending on the way the input space is modified. Their first algorithm is based on optimal subspace selection, i.e., at each boosting iteration the most relevant subset of input data is computed. The second algorithm relies on modifying the input space by means of non-linear projections. But neither method is strictly an algorithm for inducing weak classifiers from the $k$-NN rule, thus not directly addressing the problem of boosting $k$-NN classifiers. Moreover, such approaches are computationally expensive, as they rely on a genetic algorithm and a neural network, respectively.

Conversely, we propose a complete solution to the problem of boosting $k$-NN classifiers in the general multi-class setting. Namely, we propose a novel boosting algorithm, called UNN, which *induces* a *leveraged* nearest neighbors rule that generalizes the uniform $k$-NN rule. Indeed, the voting rule is redefined as a strong classifier that linearly combines weak classifiers induced by the $k$-NN rule. Therefore, our approach does not need to learn a distance function, as it directly operates on the top of $k$-nearest neighbors search. At the same time, it does not require an explicit computation of the feature space, thus preserving one of the main advantages of prototype-based methods. Our UNN boosting algorithm is an iterative procedure that learns the weights of weak classifiers, called *leveraging coefficients*. We show that this algorithm converges to the *global* minimum of any chosen *classification calibrated surrogate*[1] (Bartlett et al, 2006). Hence, our framework handles most popular losses in the machine learning literature: squared loss, exponential loss, logistic loss, etc. In particular, we prove a specific convergence rate for the exponential loss (reported in our experiments) far better than the general rate of Nock and Nielsen (2009). Another important characteristic of UNN is that it is able to discriminate the most relevant prototypes for a given class, thus allowing one for significant data reduction while improving at the same time classification performances.

1.3 Overview of the paper

In the following sections we present our approach to $k$-NN boosting. Sections 2.1-2.3 present key definitions for $k$-NN boosting. These sections also describe how to replace the classic uniform $k$-NN rule by a *leveraged* $k$-NN rule. Lever-aged $k$-NN classifiers are induced by UNN algorithm, which is detailed in Sec. 2.4 for the case of exponential risk. Sec. 2.5 presents the generic convergence theorem of UNN and the upper bound performance for the exponential risk minimization. Our experiments on both synthetic and image categorization datasets are reported in Sec. 3. Then, Sec. 4 discusses results and mentions future work.

---

[1] A *surrogate* is a function which is a suitable upperbound for another function (here, the non-convex non-differentiable empirical risk).

In order not to laden the body of the paper, the general form of UNN algorithm and proofsketches of our theorems have been postponed to an appendix in Sec. 5.

2 Method

2.1 Problem statement and notation

In this work, we address the task of *multi-class*, *single-label* image categorization. Hence, several categories of images are predefined, whereas each image is constrained to belong to a single category. The number of categories (or classes) may range from a few to hundreds, depending on applications. E.g., categorization with 67 Indoor categories has been recently studied by Quattoni and Torralba (2009). We treat the multi-class problem as multiple binary classification problems as it is customary in machine learning. I.e., for each class $c$, a query image is classified either to $c$ or to $\bar{c}$ (the complement class of $c$, which contains all classes but $c$) with a certain confidence (*classification score*). Then the label with the maximum score is assigned to the query. Images are represented by descriptors related to given local or global features. We refer to an image descriptor as an *observation* $o \in \mathcal{O}$, which is a vector of $n$ features and belongs to a *domain* $\mathcal{O}$ (e.g., $\mathbb{R}^n$ or $[0,1]^n$). A label is associated to each image descriptor according to a predefined set of $C$ classes. Hence, an observation with the corresponding label leads to an *example*, which is the ordered pair $(o, y) \in \mathcal{O} \times \mathbb{R}^C$, where $y$ is termed the *class vector* that specifies the class memberships of $o$. In particular, the sign of $y_c$ gives the membership of example $(o, y)$ to class $c$, such that $y_c$ is negative iff the observation does not belong to class $c$, positive otherwise. At the same time, the absolute value of $y_c$ may be interpreted as a relative confidence in the membership. Inspired by the multi-class boosting analysis of Zhu et al (2006), we constrain class vectors to be *symmetric*, that is:

$$\sum_{c=1}^{C} y_c = 0 \ . \tag{1}$$

Hence, in the single-label framework, the class vector of an observation $o$ belonging to class $\tilde{c}$ is defined as: $y_{\tilde{c}} = 1$, $y_{c \neq \tilde{c}} = -\frac{1}{C-1}$. This setting turns out to be necessary when treating multi-class classification as multiple binary classifications, as it balances negative and positive labels of a given example over all classes. We are given an input set of $m$ examples $\mathcal{S} = \{(\boldsymbol{o}_i, \boldsymbol{y}_i), i = 1, 2, ..., m\}$, arising from annotated images, which form the *training set*.

2.2 Boosting $k$-NN for minimization of surrogate risks

We aim at defining a one-versus-all classifier for each category, which is to be trained over the set of examples. This classifier is expected to correctly classify as many new observations as possible, *i.e.* to predict their true labels. Therefore, we aim at determining a classification rule $\boldsymbol{h}$ from the example dataset, which is able to minimize the classification error over all possible new observations. But since the underlying class probability densities are generally unknown and difficult to estimate, defining a classifier in the framework of supervised learning can be viewed as fitting a classification rule onto a training set $\mathcal{S}$ without overfitting. This corresponds to defining a classifier that correctly classifies most of the example data themselves, thus minimizing the classification error over the example dataset (empirical or true classification loss). Therefore, in the most basic framework of supervised classification, one wishes to train a *classifier* on $\mathcal{S}$, *i.e.* build a function $\boldsymbol{h} : \mathcal{O} \to \mathbb{R}^C$ with the objective to minimize its *empirical risk* on $\mathcal{S}$, defined as:

$$\varepsilon^{0/1}(\boldsymbol{h}, \mathcal{S}) \doteq \frac{1}{mC} \sum_{c=1}^{C} \sum_{i=1}^{m} [\varrho(\boldsymbol{h}, i, c) < 0] \;, \qquad (2)$$

with [.] the indicator function (1 iff true, 0 otherwise), called here the *0/1 loss*, and:

$$\varrho(\boldsymbol{h}, i, c) \doteq y_{ic} h_c(\boldsymbol{o}_i) \qquad (3)$$

the *edge* of classifier $\boldsymbol{h}$ on example $(\boldsymbol{o}_i, \boldsymbol{y}_i)$ for class $c$. Taking the sign of $h_c$ in $\{-1, +1\}$ as its membership prediction for class $c$, one sees that when the edge is positive (resp. negative), the membership predicted by classifier and the actual example's membership agree (resp. disagree). Therefore, (2) averages over all classes the number of mismatches for the membership predictions, thus measuring the goodness-of-fit of the classification rule on the training dataset. Provided that the example dataset has good generalization properties with respect to the unknown distribution of possible observations, minimizing this empirical risk is expected to yield good accuracy when classifying unlabeled observations. Unfortunately, minimizing the empirical risk is mathematically not tractable as it deals with non-convex optimization. In order to bypass this cumbersome optimization challenge, the current trend of supervised learning (including boosting and support vector machines) has replaced the minimization of the empirical risk (2) by that of a so-called *surrogate risk* (Bartlett et al, 2006), to make the optimization problem amenable. In boosting, it amounts to summing (or averaging) over classes and examples a real-valued function called the *surrogate loss*, thus ending up with the following rewriting of (2):

$$\varepsilon^{\psi}(\boldsymbol{h}, \mathcal{S}) \doteq \frac{1}{mC} \sum_{c=1}^{C} \sum_{i=1}^{m} \psi(\varrho(\boldsymbol{h}, i, c)) \;. \qquad (4)$$

Important choices available for $\psi$ include:
$$\psi^{\text{sqr}} \doteq (1-x)^2 \;, \qquad (5)$$
$$\psi^{\text{exp}} \doteq \exp(-x) \;, \qquad (6)$$
$$\psi^{\text{log}} \doteq \log(1 + \exp(-x)) \;; \qquad (7)$$

(5) is the squared loss (Bartlett et al, 2006), (6) is the exponential loss (Schapire and Singer, 1999), and (7) is the logistic loss (Bartlett et al, 2006).

*Surrogates* play a fundamental role in supervised learning. They are upper bounds of the empirical risk with desirable convexity properties. Their minimization remarkably impacts on that of the empirical risk, thus enabling to provide minimization algorithms with good generalization properties (Nock and Nielsen, 2009).

In this paper, we move from recent advances in boosting with surrogate risks to redefine the $k$-NN classification rule. In particular, we concentrate on the exponential risk and provide a novel algorithm that learns a leveraged $k$-NN classifier, while provably converging to the global optimum of a surrogate risk. Our algorithm, called UNN (Universal Nearest Neighbors), meets boosting-type convergence properties under two mild assumptions on the training set: weak learning and weak coverage properties. In the Appendix, we also describe how the UNN algorithm generalizes to any surrogate loss, and provide the most general analysis.

2.3 Leveraged $k$-NN rule

In the following, we denote by $\text{NN}_k(\boldsymbol{o}_{i'})$ the set of the $k$-nearest neighbors (with integer constant $k > 0$) of an example $(\boldsymbol{o}_{i'}, \boldsymbol{y}_{i'})$ in set $\mathcal{S}$ with respect to a non-negative real-valued "distance" function. This function is defined on domain $\mathcal{O}$ and measures how much two observations differ from each other. This dissimilarity function thus many not necessarily satisfy the triangle inequality of metrics. (All experiments in this paper refer to nearest neighbors with respect to the Euclidean distance.) For sake of readability, we let $i \sim_k i'$ denote an example $(\boldsymbol{o}_i, \boldsymbol{y}_i)$ that belongs to $\text{NN}_k(\boldsymbol{o}_{i'})$. This neighborhood relationship is intrinsically asymmetric, i.e., $i \sim_k i'$ does not necessarily imply that $i' \sim_k i$. Indeed, a nearest neighbor of $i'$ does not necessarily contain $i'$ among its own nearest neighbors.

The $k$-nearest neighbors rule ($k$-NN) is the following multi-class classifier $\boldsymbol{h} = \{h_c : c = 1, 2, ..., C\}$ ($k$ appears in the summation indices):

$$h_c(\boldsymbol{o}_{i'}) = \sum_{j \sim_k i'} [y_{jc} > 0] \;, \qquad (8)$$

where $h_c$ is the one-versus-all classifier for class $c$ and square brackets denote the indicator function. Hence, the classic nearest neighbors classification is based on majority vote among the $k$ closest prototypes.

In this paper, we propose to weight the votes of nearest neighbors by means of real coefficients, thus generalizing (8) to the following *leveraged* $k$-NN rule $\bm{h}^\ell = \{h_c^\ell : c = 1, 2, ..., C\}$:

$$h_c^\ell(\bm{o}_{i'}) = \sum_{j \sim_k i'} \alpha_{jc} y_{jc} , \qquad (9)$$

where $\alpha_{jc} \in \mathbb{R}$ is the leveraging coefficient for example $j$ in class $c$, with $j = 1, 2, ..., m$ and $c = 1, 2, ..., C$. Hence, (9) linearly combines class labels of the $k$ nearest neighbors (defined in Sec. 2.1) with their leveraging coefficients.

The main contribution of our work is to define a general algorithm (UNN) for learning these leveraging coefficients from training data. This algorithm operates on the top of classic $k$-NN methods, for it does not affect the nearest neighbors search when inducing weak classifiers of (9). Indeed, it is independent on the way nearest neighbors are computed, unlike most of the approaches mentioned in Sec. 1.2, which rely on modifying the neighborhood relationship via metric distance deformations or kernel transformations. Though, our approach is still fully compatible with any underlying (metric) distance and data structure for $k$-NN search, as well as possible kernel transformations of the input space.

For a given training set $\mathcal{S}$ of $m$ labeled examples, we define the $k$-NN *edge matrix* $\mathrm{R}^{(c)} \in \mathbb{R}^{m \times m}$ for each class $c = 1, 2, ..., C$ (Nock and Nielsen, 2009):

$$\mathrm{r}_{ij}^{(c)} \doteq \begin{cases} y_{ic} y_{jc} & \text{if } j \sim_k i \\ 0 & \text{otherwise} \end{cases} . \qquad (10)$$

The name of $\mathrm{R}^{(c)}$ is justified by an immediate parallel with (3). Indeed, each example $j$ serves as a classifier for each example $i$, predicting 0 if $j \notin \mathrm{NN}_k(\bm{o}_i)$, $y_{jc}$ otherwise, for the membership to class $c$. Hence, the $j^{th}$ column of matrix $\mathrm{R}^{(c)}$, $\bm{r}_j^{(c)}$, which is different from $\bm{0}$ when choosing $k > 0$, collects all edges of "classifier" $j$ for class $c$. Note that non-zero entries of this column correspond to the so-called *reciprocal nearest neighbors* (R$k$-NN) of $j$, i.e., those examples for which $j$ is a neighbor (Fig. 1). It finally comes that the edge of the leveraged $k$-NN rule on example $i$ for class $c$ is:

$$\varrho(\bm{h}^\ell, i, c) = (\mathrm{R}^{(c)} \bm{\alpha}^{(c)})_i , c = 1, 2, ..., C , \qquad (11)$$

where $\bm{\alpha}^{(c)}$ collects all leveraging coefficients in a vector form for class $c$: $\alpha_i^{(c)} \doteq \alpha_{ic}, i = 1, 2, ..., m$. The expression of surrogate loss (4) can be written as follows after replacing the argument of $\psi(\cdot)$ in (4) by (11):

$$\varepsilon^\psi(\bm{h}, \mathcal{S}) \doteq \frac{1}{mC} \sum_{c=1}^{C} \sum_{i=1}^{m} \psi\left( \sum_{j=1}^{m} \mathrm{r}_{ij}^{(c)} \alpha_{jc} \right) . \qquad (12)$$

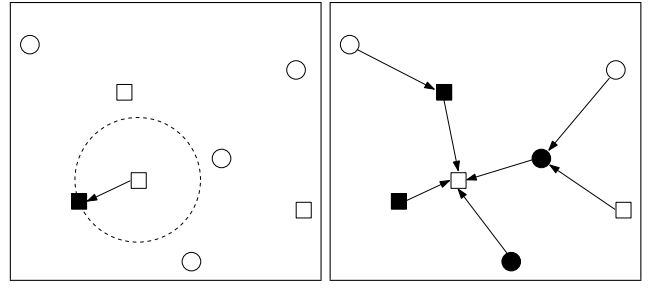

**Fig. 1** A toy example of *direct* (left) and *reciprocal* (right) $k$-nearest neighbors ($k = 1$) of an example $j$. Squares and circles represent examples of positive and negative classes. Each arrow connects an example to its 1-NN.

Therefore, fitting all $\alpha_{jc}$'s so as to minimize the surrogate loss (12) is the main goal of our learning algorithm UNN for inducing the leveraged $k$-NN classifier $\bm{h}^\ell$.

2.4 UNN: learning $\alpha_{jc}$ of leveraged $k$-NN classifier

We propose a novel classification algorithm which induces the leveraged nearest neighbors classifier $\bm{h}^\ell$ (Eq. 9) in the multi-class one-versus-all framework. In this section, we explain UNN specialized for the exponential risk minimization, with pseudo-code shown in Alg. 1. However, our analysis is much more general, as it involves the broad class of classification-calibrated surrogate risks (Bartlett et al, 2006), and is postponed to Appendix in order not to burden the methodology. Like common boosting algorithms, UNN operates on a set of weights $w_i$ ($i = 1, 2, ..., m$) defined over training data. Such weights are repeatedly updated to fit all leveraging coefficients $\bm{\alpha}^{(c)}$ for class $c$ ($c = 1, 2, ..., C$). At each iteration, the index to leverage, $j \in \{1, 2, ..., m\}$, is obtained by a call to a *weak index chooser* oracle WIC$(.,.,.)$, whose implementation is postponed to steps **[A.1]** and **[A.2]**, detailed later on in this section.

The training phase is implemented in a one-versus-all fashion, *i.e.* $C$ learning problems are solved independently, and for each class $c$ the training examples are considered as belonging to either class $c$ or the complement class $\bar{c}$, *i.e. any other class*. Eventually, one leverage coefficient ($\alpha_{jc}$) per class is learned for each weak classifier (indexed by $j$). In the Appendix, we show that Alg. 1 is a specialization of a very general classification algorithm, thus justifying the name "Universal Nearest Neighbors". In particular, Alg. 1 induces the leveraged $k$-NN classifier by minimizing the exponential surrogate risk (6), very much like regular boosting does it for inducing a weighted voting rule for a set of weak classifiers.

The key observation when training weak classifiers with UNN is that, at each iteration, one single example (indexed by $j$) is considered as a prototype to be leveraged. Indeed, all the other training data are to be viewed as observations for which $j$ may possibly vote. In particular, due to $k$-NN voting, $j$ can be a classifier only for its reciprocal nearest neighbors (i.e., those data for which $j$ itself is a neighbor, corresponding to non-zero entries in matrix (10) on column $j$). This brings to a remarkable simplification when computing $\delta_j$ in step **[I.1]** and updating weights $w_i$ in step **[I.2]** (Eq. 16, 17). Indeed, only weights of reciprocal nearest neighbors of $j$ are involved in these computations, thus allowing us not to store the entire matrix $\textsc{r}^{(c)}, c = 1, 2, ..., C$. Note that the set of R$k$-NN is splitted in two subsets, each containing examples that agree (disagree) with the class membership of $j$, thus yielding the partial sums $w_j^+$ and $w_j^-$ of (15).

Note that when whichever $w_j^+$ or $w_j^-$ is zero, $\delta_j$ in (16) is not finite. There is however a simple alternative, inspired by Schapire and Singer (1999), which consists in smoothing out $\delta_j$ when necessary, thus guaranteeing its finiteness without impairing convergence. More precisely, we suggest to replace:

$$w_j^+ \leftarrow w_j^+ + \frac{1}{m} \ , \tag{13}$$

$$w_j^- \leftarrow w_j^- + \frac{1}{m} \ . \tag{14}$$

Also note that step **[I.0]** relies on oracle $\textsc{Wic}(., ., .)$ for selecting index $j$ of the next weak classifier. We propose two alternative implementations of this oracle, as follows:

- **[I.0.a]** a lazy approach: we set $T = m$ and let $j$ be chosen by $\textsc{Wic}(\{1, 2, ..., m\}, t, c)$ either: (1) randomly, or (2) following the alphabetic order of classes;
- **[I.0.b]** the boosting approach: we pick $T \geq m$, and let $j$ be chosen by $\textsc{Wic}(\{1, 2, ..., m\}, t, c)$ such that $\delta_j$ is large enough. Each $j$ can be chosen more than once.

There are also schemes *mixing* **[I.0.a]** and **[I.0.b]**: for example, we may pick $T = m$, choose $j$ as in **[I.0.b]**, but exactly once as in **[I.0.a]**.

2.5 Properties of UNN

In this section, we enunciate two fundamental theorems for UNN. The first theorem reports a general monotonic convergence property of UNN to the optimal loss, for *any* given surrogate function. The second theorem further refines this general convergence theorem by providing effective convergence bound for the exponential loss.

**Theorem 1** *As the number of iteration steps $T$ increases, UNN converges to $h^\ell$ realizing the **global** minimum of the surrogate risk at hand (4), for any $\psi$ meeting conditions (i), (ii) and (iii) above.* (proofsketch in Appendix)

Although we prove the boosting ability of UNN for all applicable surrogate losses, we choose to show in particular its behavior for the exponential loss $\psi^{\exp}$, which features far

---

**Algorithm 1:** UNIVERSAL NEAREST NEIGHBORS UNN($\mathcal{S}$) for $\psi = \psi^{\exp}$

**Input**: $\mathcal{S} = \{(\boldsymbol{o}_i, \boldsymbol{y}_i), i = 1, 2, ..., m, \boldsymbol{o}_i \in \mathcal{O}, \boldsymbol{y}_i \in \{-\frac{1}{C-1}, 1\}^C\}$

Let $\mathrm{r}_{ij}^{(c)} \doteq \begin{cases} y_{ic}y_{jc} & \text{if } j \sim_k i \\ 0 & \text{otherwise} \end{cases}$,

$\forall i, j = 1, 2, ..., m, \ c = 1, 2, ..., C$;

**for** $c = 1, 2, ..., C$ **do**
  Let $\alpha_{jc} \leftarrow 0, \quad \forall j = 1, 2, ..., m$;
  Let $w_i \leftarrow 1, \quad \forall i = 1, 2, ..., m$;
  **for** $t = 1, 2, ..., T$ **do**
    **[I.0]** Weak index chooser oracle: Let $j \leftarrow \textsc{Wic}(\{1, 2, ..., m\}, t)$;
    **[I.1]** Let
    $$w_j^+ = \sum_{i:\mathrm{r}_{ij}^{(c)}>0} w_i, \ w_j^- = \sum_{i:\mathrm{r}_{ij}^{(c)}<0} w_i \ , \tag{15}$$
    
    $$\delta_j \leftarrow \frac{1}{2} \log\left(\frac{w_j^+}{w_j^-}\right); \tag{16}$$
    
    **[I.2]** Let
    $$w_i \leftarrow w_i \exp(-\delta_j \mathrm{r}_{ij}^{(c)}), \quad \forall i : j \sim_k i \ ; \tag{17}$$
    
    **[I.3]** Let $\alpha_{jc} \leftarrow \alpha_{jc} + \delta_j$

**Output**: $h_c(\boldsymbol{o}_{i'}) = \sum_{i \sim_k i'} \alpha_{ic} y_{ic}, \quad \forall c = 1, 2, ..., C$

---

better convergence bound than the general one (Nock and Nielsen, 2009).

Computing this bound is based on defining a *weak index assumption* (**WIA**), which is to nearest neighbors what the conventional *weak learning assumption* is to general induced classifiers (Schapire and Singer, 1999):

(**WIA**) let $p_j^{(c)} \doteq w_j^{(c)+}/(w_j^{(c)+} + w_j^{(c)-})$. There exist some $\gamma > 0$ and $\eta > 0$ such that the following two inequality holds for index $j$ returned by $\textsc{Wic}(., ., .)$:

$$|p_j^{(c)} - 1/2| \geq \gamma \ , \tag{18}$$

$$(w_j^{(c)+} + w_j^{(c)-})/\|\boldsymbol{w}\|_1 \geq \eta \ . \tag{19}$$

**Theorem 2** *If the **WIA** holds for $\tau \leq T$ steps in UNN (for each c), then $\varepsilon^{0/1}(\boldsymbol{h}^\ell, \mathcal{S}) \leq \exp(-2\eta\gamma^2\tau)$.* (proofsketch in Appendix)

Inequality (18) is the usual weak learning assumption (Schapire and Singer, 1999), when considering examples as weak classifiers. But a *weak coverage assumption* (19) is needed as well, because insufficient *coverage* of the reciprocal neighbors could easily wipe out even the surrogate risk reduction potentially due to a large $\gamma$. In addition, even when classes are significantly overlapping, choosing $k$ not too small is enough for the **WIA** to be met for a large number of boosting rounds $\tau$, thus determining a potential harsh decrease of $\varepsilon^{0/1}(\boldsymbol{h}^\ell, \mathcal{S})$. This is important, as there are at most

$m$ different weak classifiers available to WIC$(.,.,.)$, even when each one may be chosen more than once under the **WIA**. Last but not least, Theorem 2 also displays the fact that classification (18) may be more important than coverage (19).

## 3 Experiments

In this section, we present experimental results of UNN vs plain $k$-NN on both synthetic and real datasets. Such experiments allowed us to quantify the gains brought by boosting on nearest neighbors voting (Marin et al, 2009). For this purpose, we first performed tests on two-class synthetic data to drill down into the performances of UNN (Sec. 3.1). In Sec. 3.2 we discuss the data reduction ability of our technique. Then, we carried out experiments of multi-class scene categorization on a dataset of natural images and compared the results of UNN to plain $k$-NN classification (Sec. 3.3).

### 3.1 Synthetic datasets

We have drilled down into the experimental behavior of UNN using the synthetic Ripley's dataset (Ripley, 1994) with two classes denoted by P and N. Each population of this dataset is an equal mixture of two two-dimensional normally distributed populations, which are equally likely. Training and test dataset (consisting of 250 and 1000 points, respectively) are shown in Figure 2, where the optimal classification boundary of the Bayes rule is also displayed. This corresponds to the best theoretical error rate of 8.0% (Ripley, 1994).

Fig. 3 validates on this dataset the monotonous decay of the exponential risk (6), mathematically proved in Theorem 2 under the two basic weak index/learning assumptions. It also shows the effect of three different implementations of the WIC oracle (Sec. 2.5). Note that the boosting approach for selecting weak classifiers provides much faster decay of the surrogate risk, thus outperforming the two tested "lazy" implementations. In these latter cases, the index $j$ of the weak classifier at each UNN iteration was chosen either randomly or following the order of examples in their respective categories.

Classification results for a range of values of $k$ are shown in Fig. 4. They enable to draw two main conclusions: First, test errors display a robustness of UNN against variations of $k$. Second, filtering out even a large proportion $1-\theta$ of examples with the smallest $||\boldsymbol{\alpha}_.||_2^2$ does not degrade classification performances, and can even significantly improve them. As witnessed by Fig. 4, values as small as $\theta = 0.25$ yields improvements that make the test error close to Bayes'. (E.g., see the minimum error of boosted $k$-NN for $\theta = 0.25$, $k = 9$.) We investigate such a data reduction ability of UNN in the following Section.

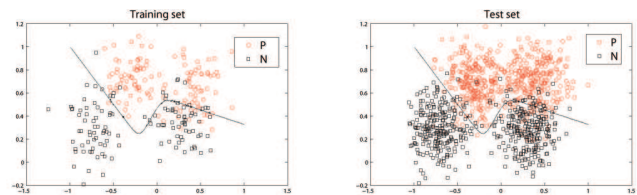

**Fig. 2** Training and validation data for the Ripley's dataset. The Bayes boundary is also drawn as reported in (Ripley, 1994).

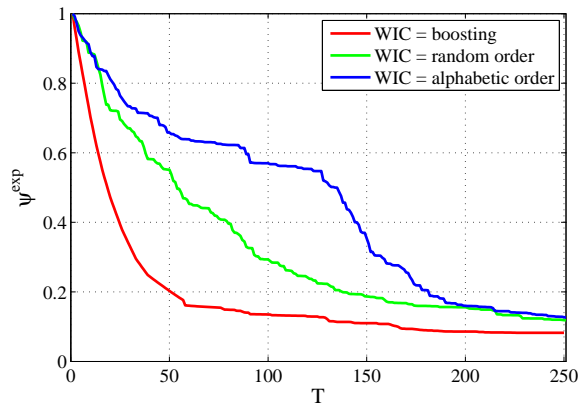

**Fig. 3** Decrease of $\varepsilon^{\psi^{\exp}}(\boldsymbol{h}^\ell, \mathcal{S})$ as a function of $T$ in UNN for the Ripley's dataset for different oracle implementations. Note that the boosting implementation (**[I.0.b]**, Sec. 2.4) always guarantees monotonic decrease of the surrogate loss, until the weak assumptions are matched (red curve). Conversely, the lazy implementation (**[I.0.a]**, Sec. 2.4) may select, at a given step, a classifier that does not match those assumptions, thus preventing the loss from strictly decreasing (see green and blue curves).

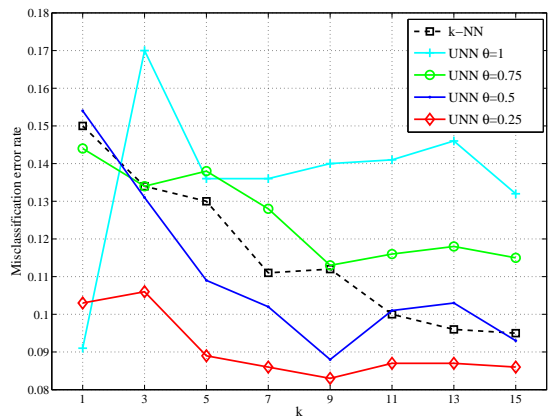

**Fig. 4** Test error for UNN as a function of $k$ for boosted $k$-NN. Bayes rule yields 8% optimal misclassification rate.

## 3.2 Filtering the prototype dataset

Experiments on the synthetic data illustrate the significant precision improvement provided by filtering the prototype dataset. Assuming standard sampling assumptions (Schapire et al, 1998), filtering benefits from two positive effects. The first is a *margin* effect, well known for *induced* classifiers (Schapire et al, 1998). The goodness-of-fit of the $k$-NN rule is driven by the most accurate examples, *i.e.* those surrounded by examples of the same class, getting the largest $||\alpha_.||_2^2$. The least accurate ones, *e.g.* those located in overlapping regions between two classes, get the smallest. Discarding these latter examples tends to increase a gap between class clouds, but each cloud may shelter examples of different classes. Fortunately, filtering with boosting is accompanied by a subtle local *repolarization* of predictions which, as explained in Figure 5 for $\theta = 0.25$, makes this gap maximization translate to *margin maximization*, for which positive effects on learning are known (Schapire et al, 1998). The second effect is structural: in nearest neighbors rules, the frontier between classes stems from the Voronoi cells of those least accurate examples. Discarding them separates better the classes, as witnessed by Fig. 5. Above all, it reduces the number of Voronoi cells involved in the class frontiers, thus reducing structural parameters (VC-dimension) of the classifier, possibly buying a reduction of the test error as well (Schapire et al, 1998).

## 3.3 Image Categorization

We tested our $k$-NN boosting algorithm for image categorization. In particular, we used the global Gist descriptor of Oliva and Torralba (2001) in order to obtain a meaningful representation of images. This descriptor provides a global representation of a scene, while not requiring explicit segmentation of image regions and objects. In the typical setting, an image is represented by a single vector of dimension 512, which collects features related to the spatial organization of dominant scales and orientations in the image. This correspondence between images and descriptors is one of the main advantages of using global descriptors over representations based on bags of local features (Grauman and Darrell, 2005). Indeed, global descriptors are straightforwardly adapted to image categorization methods relying on machine learning techniques, as most of these techniques, from prototype-based to kernel-based, require any instance of a particular category to be represented by a single vector. In particular, this is the case of $k$-NN classification, which explicitly relies on measuring one-to-one similarity between a query image and prototype images. In addition, Gist descriptors have proven successful in representing relevant contextual information of natural scenes, which allows to compute meaninfgul priors for exploration tasks like object detection and localization (Rubin et al, 2003).

The dataset we used contains 2688 color images of outdoor scenes of size 256x256 pixels, divided in 8 categories: *coast*, *mountain*, *forest*, *open country*, *street*, *inside city*, *tall buildings* and *highways*. One example image of each category is shown in Fig. 6.

To extract global descriptors from these images we used the matlab implementation by Torralba [2], with the most common settings: 4 resolution levels of the Gabor pyramid, 8 orientations per scale and $4 \times 4$ blocks.

We used this database to validate UNN for different values of $k$. In particular, we concentrated on evaluating classification performances when filtering the prototype dataset, i.e. retaining a proportion $\theta$ of the most relevant examples as prototypes for classification. Such a data reduction capability is one of the most interesting properties of UNN, as it favourably impacts on the computational cost of classification, which grows at least logarithmically (at most linearly) with the dataset size. Indeed, classification roughly amounts to searching for the $k$ nearest neighbors among prototypes, which is $O(kd\theta m)$ for linear exhaustive search, $O(kd \log(\theta m))$ for fast kD-tree based search (Arya et al, 1998) ($d$ being the dimension of feature vectors, $\theta$ the proportion of retained classifiers).

Fig. 7 shows results of 3-fold cross-validation in terms of the mean Average Precision (mAP) [3] as a function of $\theta$, for different values of $k$. Indeed, we randomly splitted the database into 3 distinct subsets, each containing 896 images. Then, for each fold, we used one of these subsets as training set, while validating on the two remaining subsets. In each experiment, UNN was run over the training set and a subset of the trained weak classifiers was retained as prototypes for classifying the test images. In particular, we selected all training images $j$ with leveraging coefficients $\alpha_{jc}$, $c = 1, 2, ..., C$, such that $\alpha_{jc} > \tilde{\alpha} > 0$. Note from Fig. 7 that, even when fixing threshold $\tilde{\alpha}$ so as to retain all the examples, the actual proportion $\theta$ of prototypes is less than one, because UNN always discards the examples with null leveraging coefficients, which do not match assumptions (18,19).

We compared UNN with the classic $k$-NN classification. Namely, in order for the classification cost of $k$-NN be roughly the same as UNN, we carried out random sampling of the prototype dataset for selecting proportion $\theta$ (between 10% and the whole set of examples). UNN significantly outperforms classic $k$-NN, even increasingly with $k$, as shown in Fig. 8(a).

---

[2] publicly available at http://people.csail.mit.edu/torralba/code/spatialenvelope/sceneRecognition.m

[3] The mAP was computed by averaging classification rates over categories (diagonal of the confusion matrix) and then averaging those values over the 3 cross-validation folds (Oliva and Torralba, 2001).

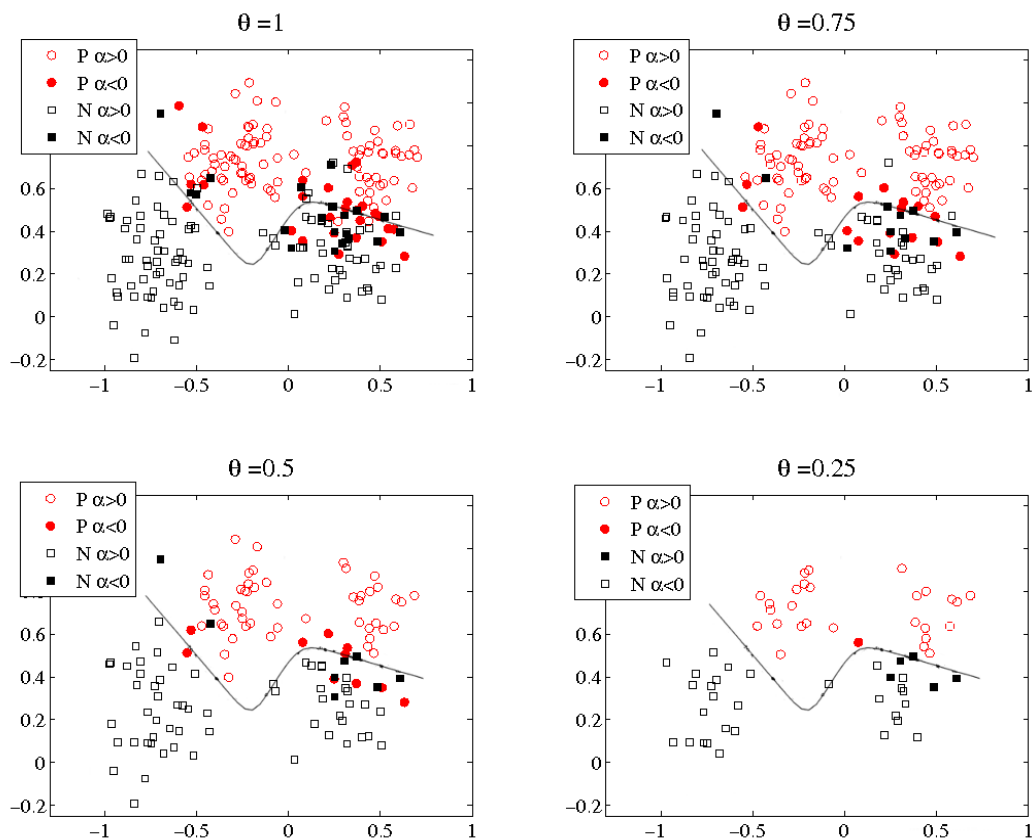

**Fig. 5** Maps of positive/negative leveraging coefficients $\alpha_j$ over training data for $k = 3$ and three different values of $\theta$. Examples of class N with negative $\alpha.$ (filled squares) and those of class P with positive $\alpha.$ (empty circles) predict class P; similarly, empty squares and filled circles both correspond to membership prediction in N. For this reason, when $\theta = 0.25$, filtering produces a clear-cut *gap* between the two possible *membership predictions* (but not between the original classes). The optimal Bayes boundary between classes is shown as well. Interestingly, while this frontier still does not separate the original classes (without error), it *does* separate the memberships predictions, with much larger minimal *margin*. The combination of the data reduction and *polarity reversal* for memberships has thus simplified the learning of $\mathcal{S}$, and eased the capture of the optimal frontier with nearest neighbors.

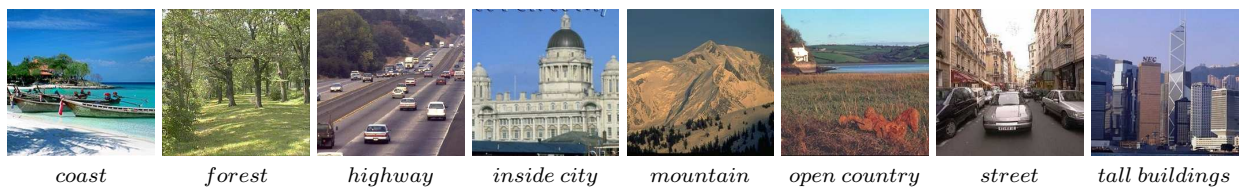

*coast*   *forest*   *highway*   *inside city*   *mountain*   *open country*   *street*   *tall buildings*

**Fig. 6** Examples of annotated images of the database containing 2688 images classified into 8 categories.

Image categorization results confirm the trend observed on the synthetic data when filtering the prototype dataset. Hence, selecting a reduced set of prototypes limits overfitting on training data, while improving classification performance on the test set (typically 3% improvement). Most interestingly, classification precision of UNN is very stable as a function of $\theta$, as it is shown in Fig. 8(b), where the drop of UNN precision for the largest values of $\theta$ is due to including prototypes with negative leveraging coefficients as well. To summarize, UNN displays the ability to discriminate the most relevant images of each class, thus inducing a classification rule robust to "noisy" prototypes arising from low inter-class variations. Adjusting the value of threshold $\tilde{\alpha}$ enables to remove those confusing prototypes, thus reducing the representation of each category to a sparse subset of meaningful prototype images.

Fig. 10 shows two examples of how the leveraged $k$-NN rule may correct misclassifications due to the uniform $k$-NN voting. E.g., in the first example, the classic and the boosted $k$-NN methods are compared when classifying an image belonging to class *coast*, with $k = 11$. The leveraged rule with as few as 20% of prototype images is able to correctly la-bel the query image (first row). Below each nearest neighbor image we show its contribution to the classifier of (9): note that negative votes are significantly smaller than positive ones (up to an order of magnitude), thus determining positive labeling with high prediction score $h_c^\ell$, according to (9). On the contrary, uniform voting rule with all prototypes misclassifies the test image, not being able to reject contributions by "noisy" neighbor images. An example of prototypes selected by filtering the dataset is shown in Fig. 11, where the leveraging coefficients refer to the first category (*tall buildings*) versus the remaining ones.

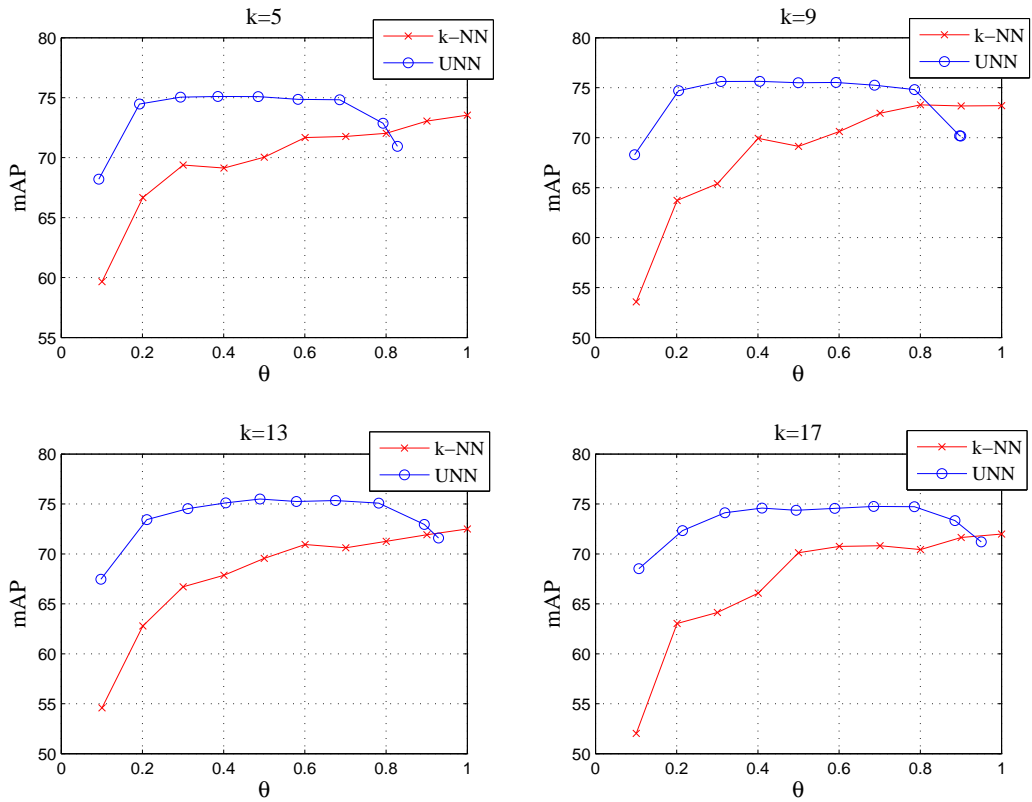

**Fig. 7** Classification performances of UNN compared to $k$-NN in 3-fold cross-validation.

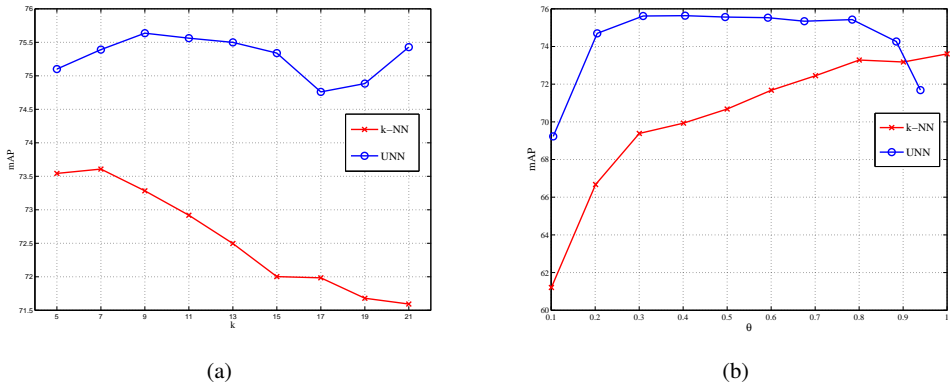

**Fig. 8** Performances of $k$-NN and UNN classification as a function of (a) $k$ and (b) $\theta$. (The best results obtained with each of the two methods are plotted.)

## 4 Conclusion

In this paper, we contribute to fill an important void of NN methods, showing how boosting can be transferred to $k$-NN classification. Namely, we propose a novel boosting algorithm, UNN (Universal Nearest Neighbors rule), for inducing a leveraged $k$-NN rule. This rule generalizes classic $k$-NN to weighted voting where weights, the so-called leveraging coefficients, are iteratively learned by UNN. We prove that this algorithm converges to the global optimum of surrogate risks under very mild assumptions.

Experiments on both synthetic and image categorization databases display that UNN provides significant performance improvements (up to the best possible performance of the Bayes rule). Moreover, UNN exhibits consistent data reduction ability, which results in significant speed-ups for classification (up to a factor 16 when removing 3/4 of the coefficients).

Our approach is built on the top of $k$-NN search, thus being fully compatible with existing techniques relying on metric distance learning (Zhang et al, 2006) as well as subspace projections like PCA (Jain, 2008) or kernel transformations of the input space, which are expected to enable significant improvements of categorization performances.

## 5 Appendix

*Generic* UNN *algorithm* The general version of UNN is shown in Alg. 2. This algorithm induces the leveraged $k$-NN rule (9) for the broad class of surrogate losses meeting conditions of Bartlett et al (2006), thus generalizing Alg. 1. Namely, we constrain $\psi$ to meet the following conditions: **(i)** $\text{im}(\psi) = \mathbb{R}_+$, **(ii)** $\nabla_\psi(0) < 0$ ($\nabla_\psi$ is the conventional derivative of $\psi$ loss function), and **(iii)** $\psi$ is strictly convex and differentiable. **(i)** and **(ii)** imply that $\psi$ is *classification-calibrated*: its local minimization is roughly tied up to that of the empirical risk (Bartlett et al, 2006). **(iii)** implies convenient algorithmic properties for the minimization of the surrogate risk (Nock and Nielsen, 2009). Three common examples have been shown in Eq. (6 – 5).

The main bottleneck of UNN is step **[I.1]**, as Eq. (21) is non-linear, *but* it always has a solution, finite under mild assumptions (Nock and Nielsen, 2009): in our case, $\delta_j$ is guaranteed to be finite when there is no total matching or mismatching of example $j$'s memberships with its reciprocal neighbors', for the class at hand. The second column of Table 1 contains the solutions to (21) for surrogate losses mentioned in Sec. 2.2. Those solutions are always exact for the exponential loss ($\psi^{\exp}$) and squared loss ($\psi^{\text{squ}}$); for the logistic loss ($\psi^{\log}$) it is exact when the weights in the reciprocal neighborhood of $j$ are the same, otherwise it is approximated. Since starting weights are all the same, exactness can be guaranteed during a large number of inner rounds depending on which order is used to choice the examples. Table 1 helps to formalize the finiteness condition on $\delta_j$ mentioned above: when either sum of weights in (20) is zero, the solutions in the first and third line of Table 1 are not finite. A simple strategy to cope with numerical problems arising from such situations is that proposed by Schapire and Singer (1999). (See Sec. 2.4.) Table 1 also shows how the weight update rule (22) specializes for the mentioned losses.

*Proofsketch of Theorem 1* We show that UNN converges to the global optimum of any surrogate risk (Sec. 2.2). So, let us consider the surrogate risk (4) for any fixed class $c =$

**Algorithm 2:** Algorithm UNIVERSAL NEAREST NEIGHBORS UNN($\mathcal{S}, \psi$)

**Input**: $\mathcal{S} = \{(\boldsymbol{o}_i, \boldsymbol{y}_i), i = 1, 2, ..., m, \boldsymbol{o}_i \in \mathcal{O}, \boldsymbol{y}_i \in \{-\frac{1}{C-1}, 1\}^C\}$, $\psi$ meeting **(i)**, **(ii)**, **(iii)** (Sec. 5);

Let $\mathrm{r}_{ij}^{(c)} \doteq \begin{cases} y_{ic} y_{jc} & \text{if } j \sim_k i \\ 0 & \text{otherwise} \end{cases}$,

$\forall i, j = 1, 2, ..., m, \ c = 1, 2, ..., C$;

**for** $c = 1, 2, ..., C$ **do**
    Let $\alpha_{jc} \leftarrow 0, \quad \forall j = 1, 2, ..., m$;
    Let $w_i \leftarrow -\nabla_\psi(0) \in \mathrm{R}_{+*}^m, \quad \forall i = 1, 2, ..., m$;
    **for** $t = 1, 2, ..., T$ **do**
        **[I.0]** Let $j \leftarrow \text{WIC}(\{1, 2, ..., m\}, t)$;
        **[I.1]** Let

$$w_j^+ = \sum_{i: \mathrm{r}_{ij}^{(c)} > 0} w_i, \quad w_j^- = \sum_{i: \mathrm{r}_{ij}^{(c)} < 0} w_i, \quad (20)$$

Let $\delta_j \in \mathbb{R}$ solution of:

$$\sum_{i=1}^m \mathrm{r}_{ij}^{(c)} \nabla_\psi \left( \delta_j \mathrm{r}_{ij}^{(c)} + \nabla_\psi^{-1}(-w_i) \right) = 0 ; \quad (21)$$

**[I.2]** $\forall i \ : \ j \sim_k i$, let

$$w_i \leftarrow -\nabla_\psi \left( \delta_j \mathrm{r}_{ij}^{(c)} + \nabla_\psi^{-1}(-w_i) \right) . \quad (22)$$

**[I.3]** Let $\alpha_{jc} \leftarrow \alpha_{jc} + \delta_j$;

**Output**: $h_c(\boldsymbol{o}_{i'}) = \sum_{i \sim_k i'} \alpha_{ic} y_{ic}, \quad \forall c = 1, 2, ..., C$

**Fig. 9** A geometric view of how UNN converges to the global optimum of (4). (See Appendix for details and notations.)

$1, 2, ..., C$:

$$\varepsilon_c^\psi(\boldsymbol{h}, \mathcal{S}) \doteq \frac{1}{m} \sum_{i=1}^m \psi(\varrho(\boldsymbol{h}, i, c)) . \quad (23)$$

Let $\boldsymbol{w}_t$ denote the $t^{th}$ weight vector inside the "**for** $c$" loop of Alg. 2 (assuming $\boldsymbol{w}_0$ is the initialization of $\boldsymbol{w}$); similarly, $\boldsymbol{h}_t^\ell$ denotes the $t^{th}$ leveraged $k$-NN rule obtained after the update in **[I.3]**. The following identity holds, whose prove follows from Nock and Nielsen (2009):

$$\psi(\varrho(\boldsymbol{h}_t^\ell, i, c)) = g + D_{\tilde{\psi}}(0\|w_{ti}) , \quad (24)$$

**Table 1** Three common loss functions and the corresponding solutions $\delta_j$ of (21) and $w_i$ of (22). (Vector $\boldsymbol{r}_j^{(c)}$ designates column $j$ of $\mathrm{R}^{(c)}$ and $||.||_1$ is the $L_1$ norm.) The rightmost column says whether it is (A)lways the solution, or whether it is when the weights of reciprocal neighbors of $j$ are the (S)ame.

| loss function | $\delta_j$ in (21) | $w_i$ in (22) | Opt |
|---|---|---|---|
| $\psi^{\exp} \doteq \exp(-x)$ | $\frac{1}{2}\log\left(\frac{w_j^{(c)+}}{w_j^{(c)-}}\right)$ | $w_i \exp\left(-\delta_j \mathrm{r}_{ij}^{(c)}\right)$ | A |
| $\psi^{\mathrm{squ}} \doteq (1-x)^2$ | $\frac{w_j^{(c)+} - w_j^{(c)-}}{2||\boldsymbol{r}_j^{(c)}||_1}$ | $w_i - 2\delta_j \mathrm{r}_{ij}^{(c)}$ | A |
| $\psi^{\log} \doteq \log(1+\exp(-x))$ | $\log\left(\frac{w_j^{(c)+}}{w_j^{(c)-}}\right)$ | $\frac{w_i \exp\left(-\delta_j \mathrm{r}_{ij}^{(c)}\right)}{1 - w_i\left(1+\exp\left(-\delta_j \mathrm{r}_{ij}^{(c)}\right)\right)}$ | S |

where $g(m) \doteq -\tilde{\psi}(0)$ does not depend on the $k$-NN rule. Eq. (24) makes the connection between the real-valued classification problem and a geometric problem in the non-metric space of weights. Here, we have made use of the following notations: $\tilde{\psi}(x) \doteq \psi^\star(-x)$, where $\psi^\star(x) \doteq x\nabla_\psi^{-1}(x) - \psi(\nabla_\psi^{-1}(x))$ is the Legendre conjugate of $\psi$; $D_{\tilde{\psi}}(w_i||w_i') \doteq \tilde{\psi}(w_i) - \tilde{\psi}(w_i') - (w_i - w_i')\nabla_{\tilde{\psi}}(w_i')$ is the Bregman divergence with generator $\tilde{\psi}$ (Nock and Nielsen, 2009). $\psi^\star$ is related to $\psi$ in such a way that $\nabla_{\tilde{\psi}}(x) = -\nabla_\psi^{-1}(-x)$. Eq. (24) proves in handy as one computes the *difference* $\varepsilon_c^\psi(\boldsymbol{h}_{t+1}^\ell, \mathcal{S}) - \varepsilon_c^\psi(\boldsymbol{h}_t^\ell, \mathcal{S})$. Indeed, using (24) in (23), and computing $\delta_j$ in (21) so as to bring $\boldsymbol{h}_{t+1}^\ell$ from $\boldsymbol{h}_t^\ell$, we obtain:

$$\varepsilon_c^\psi(\boldsymbol{h}_{t+1}^\ell, \mathcal{S}) - \varepsilon_c^\psi(\boldsymbol{h}_t^\ell, \mathcal{S}) = -\frac{1}{m}\sum_{i=1}^m D_{\tilde{\psi}}\left(w_{(t+1)i}||w_{ti}\right). \tag{25}$$

Since Bregman divergences are non negative and meet the identity of the indiscernibles, (25) implies that steps **[I.1]** — **[I.3]** *guarantee* the decrease of (23) as long as $\delta_j \neq 0$. But (23) is lowerbounded, hence UNN must converge. In addition, it converges to the global optimum of (23). Since predictions for each class are independent, the prove consists in showing that (23) converges to its global minimum for each $c$. Assume this convergence for the current class, $c$. Then, following Nock and Nielsen (2009), (21) and (22) imply that, when any possible $\delta_j = 0$, the weight vector, say $\boldsymbol{w}_\infty$, satisfies $\mathrm{R}^{(c)\top}\boldsymbol{w}^\top = \boldsymbol{0}$, *i.e.*, $\boldsymbol{w}_\infty \in \ker \mathrm{R}^{(c)\top}$, and $\boldsymbol{w}_\infty$ is unique. But the kernel of $\mathrm{R}^{(c)\top}$ and $\overline{\mathbb{W}}$, the closure of $\mathbb{W}$, are provably Bregman orthogonal (Nock and Nielsen, 2009), thus yielding:

$$\underbrace{\sum_{i=1}^m D_{\tilde{\psi}}(0||w_i)}_{m\varepsilon_c^\psi(\boldsymbol{h}^\ell,\mathcal{S})-mg} = \underbrace{\sum_{i=1}^m D_{\tilde{\psi}}(0||w_{\infty i})}_{m\varepsilon_c^\psi(\boldsymbol{h}_\infty^\ell,\mathcal{S})-mg}$$
$$+ \underbrace{\sum_{i=1}^m D_{\tilde{\psi}}(w_{\infty i}||w_i)}_{\geq 0}, \forall \boldsymbol{w} \in \overline{\mathbb{W}}. \tag{26}$$

Underbraces use (24) in (23), and $\boldsymbol{h}^\ell$ is a leveraged $k$-NN rule corresponding to $\boldsymbol{w}$. One obtains that $\boldsymbol{h}_\infty^\ell$ achieves the global minimum of (23), as claimed.

The proofsketch is graphically summarized in Figure 9. In particular, two crucial *Bregman orthogonalities* are mentioned (Nock and Nielsen, 2009). The red one symbolizes:

$$\sum_{i=1}^m D_{\tilde{\psi}}(0||w_{ti}) = \sum_{i=1}^m D_{\tilde{\psi}}(0||w_{(t+1)i}) + \sum_{i=1}^m D_{\tilde{\psi}}(w_{(t+1)i}||w_{ti}), \tag{27}$$

which is equivalent to (25). The black one on $\boldsymbol{w}_\infty$ is (26).

*Proofsketch of Theorem 2* Using developments analogous to those of Nock and Nielsen (2009), UNN can be shown to be equivalent to AdaBoost in which $m$ weak classifiers are available, each one being an example. Each weak classifier returns a value in $\{-1, 0, 1\}$, where 0 is reserved for examples outside the reciprocal neighborhood. Theorem 3 of Schapire and Singer (1999) brings in our case:

$$\varepsilon^{0/1}(\boldsymbol{h}^\ell, \mathcal{S}) \leq \frac{1}{C}\sum_{c=1}^C \prod_{t=1}^T Z_t^{(c)}, \tag{28}$$

where $Z_t^{(c)} \doteq \sum_{i=1}^m \tilde{w}_{it}^{(c)}$ is the normalizing coefficient for each weight vector in UNN. ($\tilde{w}_{it}^{(c)}$ denotes the weight of example $i$ at iteration $(t,c)$ of UNN, and the Tilda notation refers to weights normalized to unity at each step.) It follows that:

$$Z_t^{(c)} = 1 - \tilde{w}_{jt}^{(c)+-}\left(1 - 2\sqrt{p_{jt}^{(c)}(1-p_{jt}^{(c)})}\right)$$
$$\leq \exp\left(-\tilde{w}_{jt}^{(c)+-}\left(1 - 2\sqrt{p_{jt}^{(c)}(1-p_{jt}^{(c)})}\right)\right)$$
$$\leq \exp\left(-\eta\left(1 - \sqrt{1-4\gamma^2}\right)\right) \leq \exp(-2\eta\gamma^2),$$

where $\tilde{w}_{jt}^{(c)+-} \doteq \tilde{w}_{jt}^{(c)+} + \tilde{w}_{jt}^{(c)-}$, $p_{jt}^{(c)} \doteq \tilde{w}_{jt}^{(c)+}/\tilde{w}_{jt}^{(c)+-} = w_{jt}^{(c)+}/w_{jt}^{(c)+-}$. The first inequality uses $1 - x \leq \exp(-x)$, and the second the **WIA**. Since even when the **WIA** does not hold, we still observe $Z_t^{(c)} \leq 1$, plugging the last inequality in (28) yields the statement of the Theorem.

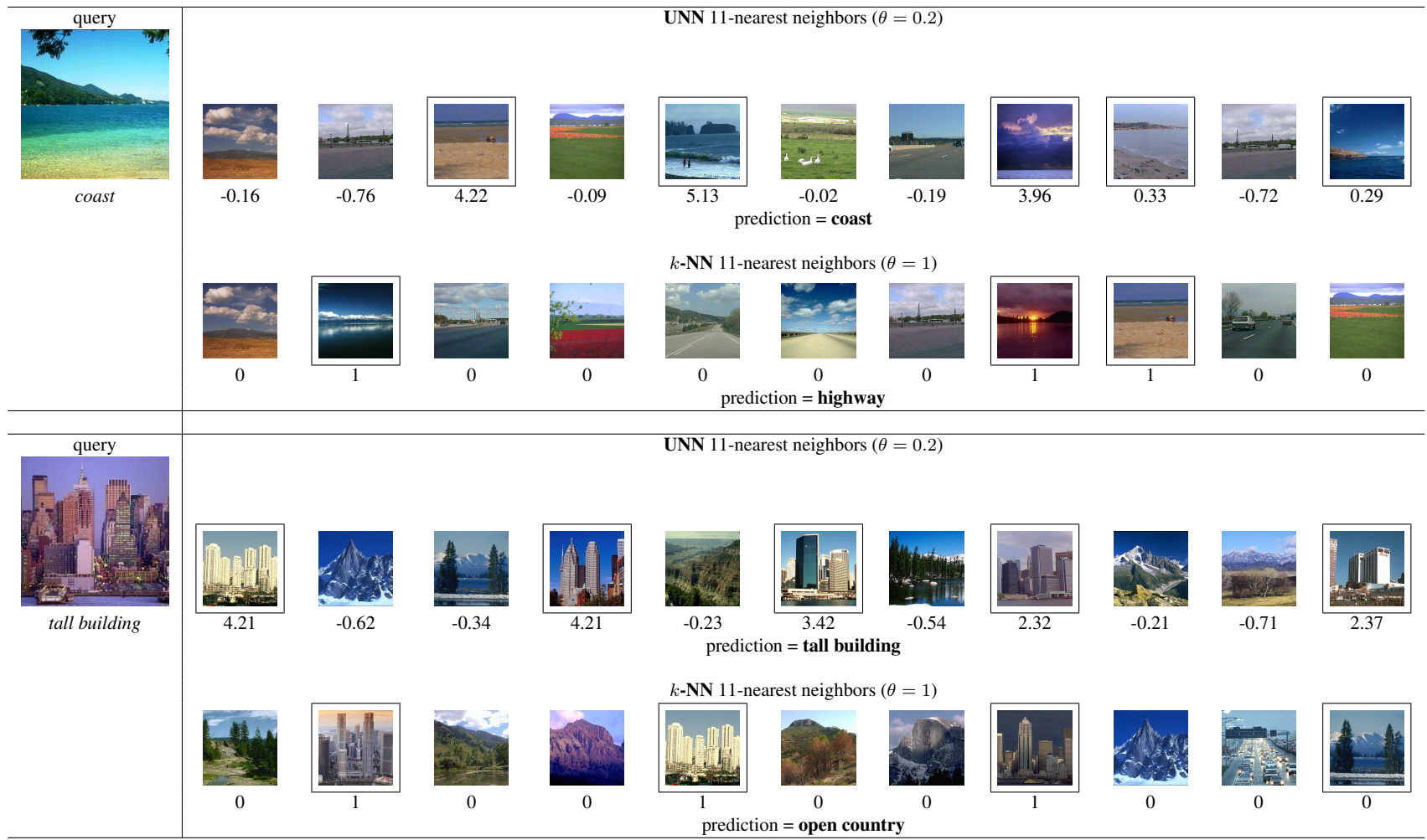

**Fig. 10** Two examples where UNN corrects misclassifications of $k$-NN. The query image is shown in the leftmost column. The 11-nearest prototype images are shown on the right: the first row refers to UNN with 20% of retained prototypes ($\theta = 0.2$), whereas the second column refers to classic $k$-NN classification over all prototypes ($\theta = 1$). Neighbors in the same category as the query image are surrounded by black boxes. Votes given by each prototype for the true category (*coast*) are shown below each image (such values correspond to $\alpha_{ic} y_{jc}$ in (9), where $c$ is the ground-truth category).

| tall buildings | inside city | street | highway | coast | forest | mountain | open country |
|---|---|---|---|---|---|---|---|
| 3.41 | 1.87 | 1.33 | 1.33 | 3.78 | 3.75 | 4.38 | 3.45 |
| 4.37 | 2.32 | 2.75 | 1.66 | 3.41 | 2.32 | 4.84 | 3.43 |
| 1.74 | 2.72 | 3.41 | 1.97 | 1.43 | 3.41 | 1.69 | 5.35 |
| 3.40 | 3.43 | 3.97 | 4.44 | 4.22 | 4.79 | 1.44 | 1.51 |
| 3.42 | 5.01 | 1.54 | 5.31 | 3.46 | 1.33 | 4.55 | 1.89 |
| 1.34 | 1.37 | 3.77 | 1.87 | 4.64 | 1.59 | 3.40 | 4.30 |
| 3.40 | 4.68 | 1.73 | 5.05 | 3.44 | 2.40 | 1.90 | 2.02 |
| 2.25 | 1.70 | 3.76 | 1.45 | 5.13 | 1.81 | 2.05 | 4.44 |
| 3.32 | 2.45 | 3.45 | 1.59 | 3.43 | 1.50 | 3.76 | 1.49 |
| 4.21 | 3.42 | 1.35 | 2.38 | 1.52 | 4.10 | 5.26 | 4.92 |
| 3.95 | 3.76 | 5.30 | 3.43 | 4.82 | 1.37 | 1.59 | 4.10 |
| 1.73 | 2.03 | 3.96 | 4.30 | 1.95 | 1.79 | 1.73 | 1.98 |
| 4.10 | 3.96 | 4.44 | 5.30 | 4.10 | 3.77 | 3.75 | 4.51 |
| 1.62 | 1.84 | 4.79 | 2.14 | 2.12 | 3.45 | 1.49 | 1.96 |
| 3.75 | 1.66 | 1.78 | | 3.96 | 2.01 | 2.16 | 1.86 |
| 2.32 | 3.96 | 1.65 | | 3.41 | 4.95 | 4.37 | 1.93 |
| 1.31 | 1.78 | 4.44 | | 4.55 | 4.10 | 4.55 | 4.21 |
| 1.59 | 1.37 | 4.09 | | 2.01 | 1.65 | 2.23 | 2.05 |
| 3.95 | 3.41 | 2.02 | | 1.92 | 4.87 | 1.86 | 5.13 |
| 1.35 | 4.21 | | | | 4.44 | 3.41 | 3.97 |
| 5.03 | 1.40 | | | | 3.77 | 1.83 | |
| 1.38 | 3.96 | | | | 4.50 | 1.88 | |
| 3.75 | 4.50 | | | | 1.55 | 3.41 | |
| 3.42 | 2.57 | | | | 3.41 | | |

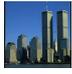

**Fig. 11** Examples of image prototypes with their leveraging coefficients for category 1 (*tall buildings*)-versus-all.